\documentclass[conference]{IEEEtran}
\IEEEoverridecommandlockouts
\usepackage{cite}
\usepackage[colorlinks,
            linkcolor=red,
            anchorcolor=green,
            citecolor=blue
            ]{hyperref}
\usepackage{amsmath,amssymb,amsfonts}
\usepackage{algorithmic}
\usepackage{graphicx}
\usepackage{textcomp}
\usepackage{xcolor}
\usepackage{xcolor}
\usepackage{siunitx}   
\usepackage{booktabs}  
\usepackage{multirow} 
\usepackage{xcolor}
\usepackage{float} 
\usepackage[table]{xcolor}
\usepackage{placeins}
\usepackage{subcaption}
\usepackage{url}
\usepackage{bm}
\def\BibTeX{{\rm B\kern-.05em{\sc i\kern-.025em b}\kern-.08em
    T\kern-.1667em\lower.7ex\hbox{E}\kern-.125emX}}
\begin{document}

\title{VeloGauss: Learning Physically Consistent Gaussian Velocity Fields from Videos}

\author{
    \IEEEauthorblockN{
    Nengbo  Lu\textsuperscript{},
    Bin Zhao\textsuperscript{*}
    }
    \IEEEauthorblockA{
    \textsuperscript{} Guangxi Key Laboratory of Robot Intelligent Perception and Control, \\ School of Artificial Intelligence, Guilin University of Electronic Technology, Guilin, 541004, China \\
zhaobinnku@mail.nankai.edu.cn
    }
}
\renewcommand{\thefootnote}{\fnsymbol{footnote}}

\maketitle
\footnotetext[1]{Corresponding author.}
\footnotetext{This work was supported by NSFC under 62566017}

\maketitle

\begin{abstract}
In this paper, we aim to jointly model the geometry, appearance, and physical information of 3D scenes solely from dynamic multi-view videos, without relying on any physical priors. Existing works typically employ physical losses merely as soft constraints or integrate physical simulations into neural networks; however, these approaches often fail to effectively learn complex motion physics. Although modeling velocity fields holds the potential to capture authentic physical information, due to the lack of appropriate physical constraints, current methods are unable to correctly learn the interaction mechanisms between rigid and non-rigid particles. To address this, we propose VeloGauss, designed to learn the physical properties of complex dynamic 3D scenes without physical priors. Our method learns the velocity field for each Gaussian particle by introducing a Physics Code and a Particle Dynamics System, and ultimately incorporates Global Physical Constraints to ensure the physical consistency of the scene. Extensive experiments on four public datasets demonstrate that our method outperforms achieves state-of-the-art performance in both Novel View Interpolation and Future Frame Extrapolation tasks. To adhere to the principles of Double-Blind
Review, we anonymously provide the code at \url{https://github.com/DustSettled/VeloGauss}
\end{abstract}


\section{Introduction}
Modeling dynamic 3D scenes from monocular or multi-view videos remains a central pursuit in computer vision. In recent years, the emergence of Neural Radiance Fields (NeRF)\cite{mildenhall2020nerfrepresentingscenesneural} and 3D Gaussian Splatting (3DGS)\cite{kerbl20233dgaussiansplattingrealtime} has significantly enriched the spectrum of representations available for dynamic scene reconstruction. Initially, NeRF leveraged implicit neural networks and Volumetric Rendering to enable continuous scene representation and high-fidelity novel view synthesis. Subsequently, 3DGS introduced an explicit representation based on 3D Gaussian ellipsoids. By utilizing efficient Differentiable Rasterization to replace computationally expensive ray marching, 3DGS has achieved real-time rendering speeds while maintaining photorealistic quality.

\begin{figure}
    \centering
    \includegraphics[width=0.5\textwidth,height=6cm]{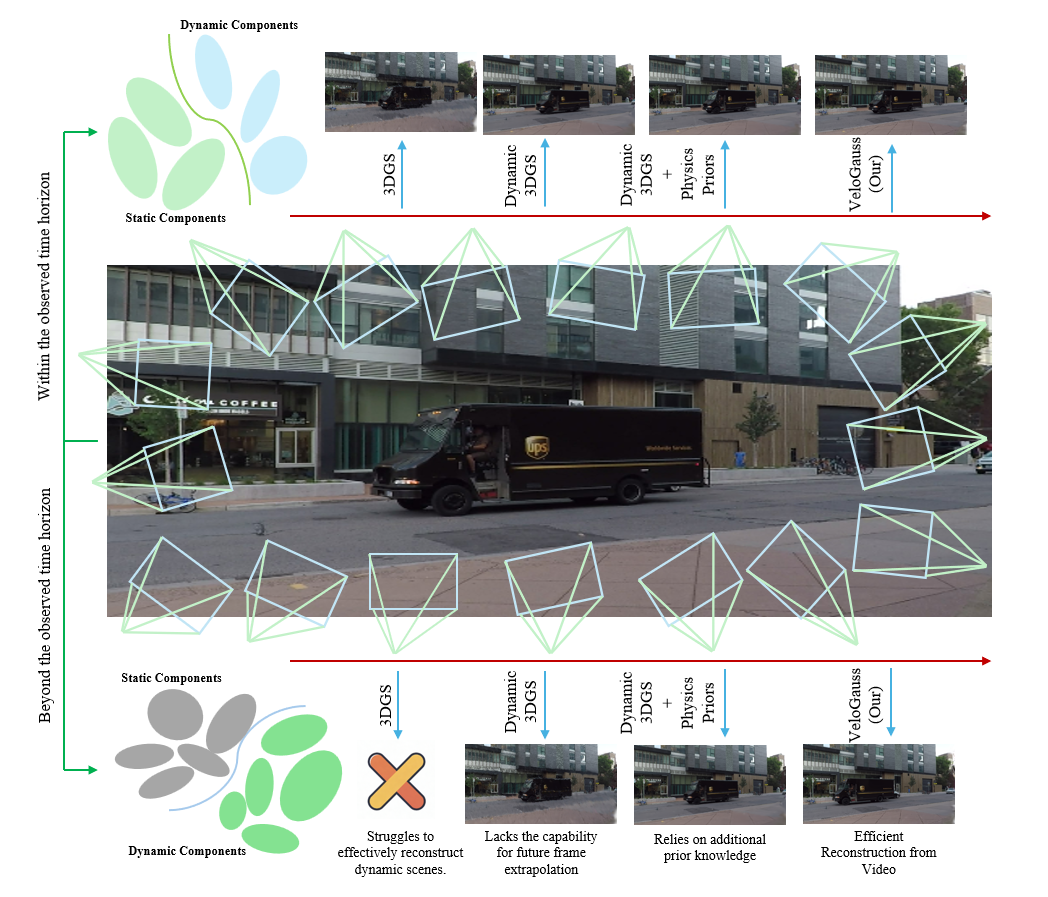}
    \caption{Our approach achieves superior efficiency and significantly outperforms existing methods\cite{kerbl20233dgaussiansplattingrealtime,yang2024deformable}, both within and beyond the observed time horizon.}
    \vspace{-0.5cm}
    \label{fig:frame2}
\end{figure}

However, extending this success from static to dynamic scenes still poses severe challenges. To address object motion, existing dynamic 3DGS methods typically employ Temporal Latent Codes\cite{duan20244drotorgaussiansplattingefficient,li2024spacetimegaussianfeaturesplatting} to modulate Gaussian attributes or utilize continuous Explicit Deformation Fields\cite{huang2024scgssparsecontrolledgaussiansplatting,gao2024gaussianflowsplattinggaussiandynamics,wu20244dgaussiansplattingrealtime,liang2025gaufregaussiandeformationfields} to fit scene variations. While these approaches perform remarkably well within the observed time horizon, achieving high-quality modeling results, they fundamentally amount to geometric overfitting of the observed data. The lack of understanding regarding underlying physical mechanisms leads to poor generalization capabilities on unseen timestamps. Consequently, it becomes difficult to guarantee that motion trajectories adhere to basic physical conservation laws, thereby precluding reliable prediction of future frames, as shown in Figure \ref{fig:frame2}.

To understand underlying physical mechanisms and learn intrinsic physical information, existing research primarily falls into two categories. The first involves the explicit integration of physics models\cite{xie2024physgaussianphysicsintegrated3dgaussians,borycki2025gaspgaussiansplattingphysicbased,kaneko2024improvingphysicsaugmentedcontinuumneural}, attempting to couple 3DGS with traditional solvers such as the Material Point Method (MPM). While physically rigorous, these methods typically require laborious manual tuning of material parameters and precise geometric initializations\cite{Huang_2024}. Furthermore, they struggle to infer unknown physical properties directly from raw video inputs. The second category employs Physics-Informed Neural Networks (PINNs)\cite{Chu_2022} to integrate Partial Differential Equations (PDEs) into the learning objective\cite{hong2025physicsinformeddeformablegaussiansplatting}. However, the direct application of traditional PINNs faces numerous challenges. They are generally difficult to optimize and computationally inefficient due to the calculation of high-order derivatives. Moreover, they are prone to over-smoothing, where physical constraints conflict with visual reconstruction losses during early training, leading to issues such as blurred boundaries. Additionally, these methods often rely on strong priors\cite{zhang2024physdreamerphysicsbasedinteraction3d,zhu2024motiongsexploringexplicitmotion}, such as object masks, which significantly limits their generalizability to unconstrained videos.

\begin{figure*}
    \centering
    \includegraphics[width=1\textwidth, height=4.5cm]{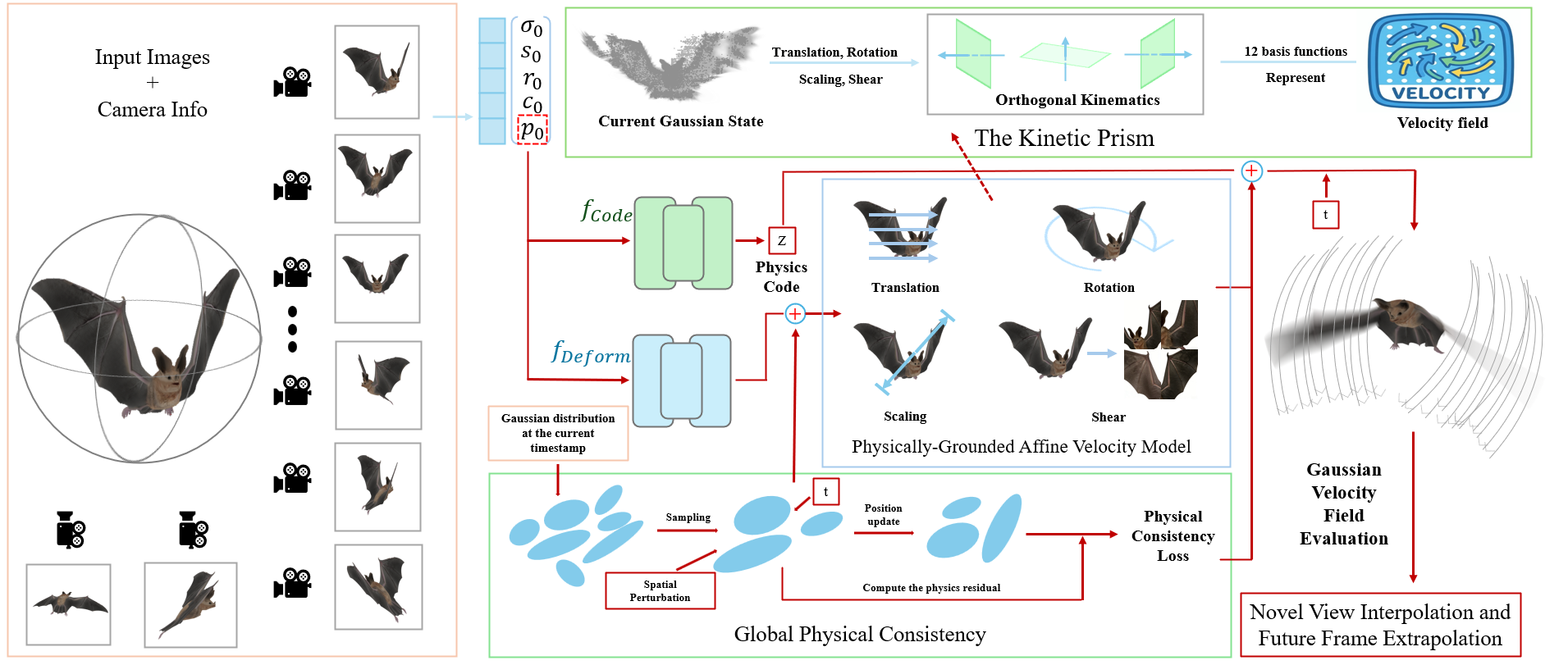}
    \caption{Taking multi-view RGB video streams as input, our method first employs a physics encoder to identify per-particle latent physics codes. Driven by these codes, the Physically-Grounded Affine Velocity Model explicitly factorizes complex hybrid dynamics into a representation spanned by 12 physical basis functions. The velocity field learning is rigorously regularized by Global Physical Consistency, while an Auxiliary Deformation Field is incorporated to capture residual high-frequency geometric details. }
    \vspace{-0.5cm}
    \label{fig:frame0}
\end{figure*}

In this paper, we introduce a novel framework designed to model complex dynamic 3D scenes and parse their physical characteristics\cite{liu2024physics3dlearningphysicalproperties,xu2024gaussianpropertyintegratingphysicalproperties} solely from multi-view RGB videos, without relying on any manual annotations. Among various physical attributes, we establish Velocity as the core learnable primitive. Serving as a natural bridge connecting instantaneous geometry with underlying physical mechanisms, the velocity field offers a unique dual advantage: it not only enables physically consistent Future Extrapolation\cite{li2025freegave,li2023nvfi} via time integration but also allows for direct regularization by physical laws.

\begin{figure}
    \centering
    \includegraphics[width=0.5\textwidth]{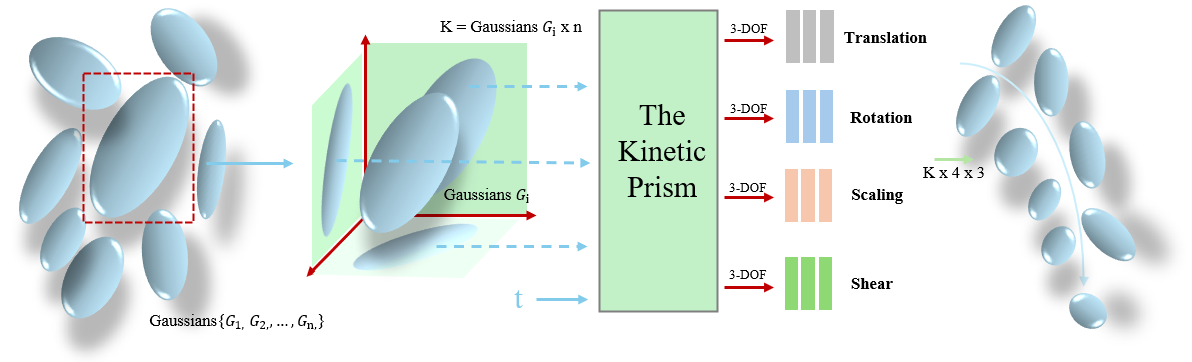}
    \caption{Network framework for constructing the Gaussian velocity field by learning basic velocity components.}
    \vspace{-0.5cm}
    \label{fig:frame}
\end{figure}

However, accurately learning physically compliant velocity solely from RGB videos is extremely challenging, fundamentally because Gaussian primitives in the raw 3D space lack sufficient physical constraints\cite{mo2025plugandplaypdeoptimization3d,wang2025odegslatentodesdynamic}. This difficulty is further exacerbated by complex motion discrepancies among multiple objects. Based on this insight, we propose to construct an independent dynamics system for every particle, encompassing both rigid and non-rigid types. Following the laws of classical mechanics, we decompose particle motion into self-centric translation, rotation, scaling, and shear. By learning this comprehensive set of dynamic parameters for each particle, we can construct a velocity field that strictly adheres to the scene's physical rules. Furthermore, we argue that the limitations of PINNs in 3DGS applications stem not from the physical equations themselves, but from the representation and optimization strategies employed. To this end, our framework introduces a robust learning strategy that innovatively unifies Lagrangian particle tracking with Eulerian velocity fields, thereby imposing effective and robust physical constraints on dynamic scene reconstruction\cite{wu20244dgaussiansplattingrealtime,wu2025deblur4dgs4dgaussiansplatting,zhang2025megamemoryefficient4dgaussian}.

As shown in Figure \ref{fig:frame0}, our framework consists of three synergistic core components:
1) 3D Scene Representation Module: Built upon 3DGS, this module is responsible for efficiently representing the dynamic scene geometry and appearance details within the observed time horizon;
2) Particle Dynamics System Module: Designed to estimate a complete set of physical parameters for each input particle. Utilizing a Velocity Decomposition strategy, this module employs Multilayer Perceptrons (MLPs) to predict instantaneous particle motion states and derives a physically compliant velocity field in accordance with the laws of classical mechanics;
3) Physical Constraint Module: Dedicated to enforcing global physical consistency during the learning process. This is realized through an optimized and re-engineered PINN, ensuring that the generated motion trajectories adhere to physical conservation laws.

To mitigate the inaccuracy and instability of Gaussian kernel regression during early training stages, we strategically introduce an Auxiliary Deformation Field within our second module for parallel optimization. Distinct from prevailing dynamic approaches based on NeRF and 3DGS, our core innovation lies in the deep integration of a Particle Dynamics System and a Physical Constraint Module. This design enables the model to go beyond simple geometric fitting and truly learn the underlying physical mechanisms of the scene, achieving significant performance gains in both interpolation reconstruction and Future Frame Extrapolation.
We name our method VeloGauss, which leverages 3D Gaussians as a foundation to accurately model scene geometry and appearance, resolves complex motion relationships by learning a microscopic particle dynamics system, and incorporates global physical constraints to optimize the entire learning process.
Our main contributions are summarized as follows:
\begin{itemize}
    \item We propose a physics-aware general framework capable of modeling complex dynamic 3D scenes solely from RGB videos, without requiring any manual annotations or prior physical knowledge.
    \item We construct a per-particle dynamics system that accurately characterizes the motion parameters of each particle, allowing for the bottom-up derivation of a continuous and physically compliant velocity field based on classical mechanics.
    \item We design a physical constraint optimization mechanism that utilizes governing equations as regularization terms. This effectively suppresses non-physical artifacts and empowers the model with reliable capabilities for future frame prediction.
\end{itemize}

\section{method}
\subsection{Canonical 3D Representation}
To effectively decouple geometry from motion, we establish a Canonical Space defined at the reference timestamp $t=0$. Unlike implicit neural representations, we adopt the explicit 3DGS as the underlying scene representation. This particle-based representation naturally aligns with our Lagrangian perspective, where each Gaussian primitive is treated as an independent physical entity. 

Formally, the canonical scene consists of a set of 3D Gaussian primitives $\mathcal{G} = \{G_1, \dots, G_N\}$. Each particle $G_i$ is parameterized by a set of learnable attributes: center position $\bm{\mu}_i \in \mathbb{R}^3$ in the world coordinate system, anisotropic scaling factor $\mathbf{s}_i \in \mathbb{R}^3$, rotation quaternion $\mathbf{q}_i \in \mathbb{R}^4$ representing spatial pose, opacity $\alpha_i \in [0, 1]$, and Spherical Harmonic (SH) coefficients $\mathbf{c}_i$ for view-dependent color modeling. To introduce rigid body dynamics constraints, we regard each Gaussian kernel as a \textbf{rigid particle} with non-zero volume. Its covariance matrix $\mathbf{\Sigma}_i$ can be expressed as
\begin{equation}
    \mathbf{\Sigma}_i = \mathbf{R}_i \mathbf{S}_i \mathbf{S}_i^T \mathbf{R}_i^T,
    \label{eq:covariance}
\end{equation}
where $\mathbf{R}_i$ is the rotation matrix derived from $\mathbf{q}_i$, and $\mathbf{S}_i$ is the diagonal scaling matrix derived from $\mathbf{s}_i$.

During the rendering phase, the deformed Gaussian particles are projected onto the 2D image plane via differentiable rasterization. The final pixel color $C$ is synthesized using the standard $\alpha$-blending formula
\begin{equation}
    C = \sum_{i \in \mathcal{N}} \mathbf{c}_i \alpha_i \prod_{j=1}^{i-1} (1 - \alpha_j),
    \label{eq:rendering}
\end{equation}
where $\mathcal{N}$ denotes the set of ordered Gaussians overlapping the pixel.

\subsection{Particle Dynamics System}
This module is designed to precisely identify the physical parameters governing each Gaussian primitive, thereby enabling accurate modeling of the microscopic Gaussian velocity field. Real-world dynamic 3D scenes exhibit immense spatiotemporal complexity, where the evolution typically manifests as a hybrid dynamic process characterized by the intertwining of rigid body motions and non-rigid deformations. Given that directly modeling such a globally coupled field is prohibitively challenging, we focus on decoupling and learning the kinematic characteristics of individual Gaussian particles. This design strategy significantly reduces optimization complexity, rendering the learning of physical mechanisms in complex scenes highly efficient.

To effectively mitigate the complexity of learning high-dimensional velocity fields while strictly enforcing physical plausibility, we propose a \textbf{Physically-Grounded Affine Velocity Model}, as shown in Figure \ref{fig:frame}. Distinct from conventional paradigms that utilize black-box networks to directly regress per-point displacements, we adopt a \textbf{Spatiotemporal Decoupling} strategy. This approach decomposes complex dynamics into \textit{explicit spatial geometric structures} and \textit{implicit temporal evolution coefficients}. We formulate the instantaneous motion of any local region within the dynamic scene as a 12-degree-of-freedom affine transformation. Accordingly, the velocity field $\mathbf{v}(\mathbf{x}, t)$ is parameterized as a time-varying linear combination of a set of fixed physical basis functions. We define a spatial basis function bank $\bm{\Phi}(\mathbf{x}) = [\bm{\phi}_1(\mathbf{x}), \dots, \bm{\phi}_{12}(\mathbf{x})]$, where $\mathbf{x} \in \mathbb{R}^3$ denotes the particle's canonical position. These bases comprehensively cover all degrees of freedom for 3D affine transformations. For any arbitrary spatial coordinate $\mathbf{x} = [x, y, z]^T$, these basis functions are explicitly constructed as:
\begin{equation}
    \bm{\Phi}(\mathbf{x}) = \underbrace{\left[ \mathbf{I}_3 \right]}_{\text{Translation}} \oplus \underbrace{\left[ \begin{smallmatrix} 0 & z & -y \\ -z & 0 & x \\ y & -x & 0 \end{smallmatrix} \right]}_{\text{Rotation}} \oplus \underbrace{\left[ \text{diag}(\mathbf{x}) \right]}_{\text{Stretch}} \oplus \underbrace{\left[ \begin{smallmatrix} y & 0 & z \\ x & z & 0 \\ 0 & y & x \end{smallmatrix} \right]}_{\text{Shear}},
    \label{eq:basis_definition}
\end{equation}
where $\mathbf{I}_3$ denotes the identity matrix, and $\oplus$ represents the column-wise concatenation operation. As shown in Figure \ref{fig:frame}, this explicit construction affords interpretability, ensuring that the learned velocity field inherently adheres to geometric constraints, thereby preventing non-physical distortions commonly observed in unstructured implicit fields.

\begin{table*}[htbp]
\centering
\caption{Quantitative evaluation of novel view interpolation and future frame extrapolation on the Dynamic Object and Dynamic Indoor Scene datasets\cite{li2023nvfi}.}
\label{tab:dynamic_datasets}
\resizebox{\linewidth}{!}{
\label{tab:table1}
\begin{tabular}{l|ccc|ccc|ccc|ccc}
\toprule
\multirow{3}{*}{Method} & \multicolumn{6}{c|}{Dynamic Object Dataset} & \multicolumn{6}{c}{Dynamic Indoor Scene Dataset} \\
\cmidrule{2-13}
& \multicolumn{3}{c|}{Interpolation} & \multicolumn{3}{c|}{Extrapolation} & \multicolumn{3}{c|}{Interpolation} & \multicolumn{3}{c}{Extrapolation} \\
\cmidrule{2-4} \cmidrule{5-7} \cmidrule{8-10} \cmidrule{11-13}
& PSNR$\uparrow$ & SSIM$\uparrow$ & LPIPS$\downarrow$ & PSNR$\uparrow$ & SSIM$\uparrow$ & LPIPS$\downarrow$ & PSNR$\uparrow$ & SSIM$\uparrow$ & LPIPS$\downarrow$ & PSNR$\uparrow$ & SSIM$\uparrow$ & LPIPS$\downarrow$ \\
\midrule
T-NeRF{\cite{pumarola2021d}}       & 13.163 & 0.709 & 0.353 & 13.818 & 0.739 & 0.324 & 24.944 & 0.742 & 0.336 & 22.242 & 0.700 & 0.363 \\
D-NeRF{\cite{pumarola2021d}}       & 14.158 & 0.697 & 0.352 & 14.660 & 0.737 & 0.312 & 25.380 & 0.746 & 0.300 & 20.791 & 0.692 & 0.369 \\
T-NeRF$_{\text{PINN}}$ & 15.286 & 0.794 & 0.293 & 16.189 & 0.835 & 0.230 & 26.250 & 0.461 & 0.638 & 23.290 & 0.477 & 0.414 \\
HexPlane$_{\text{PINN}}${\cite{cao2023hexplane}}  & 27.042 & 0.958 & 0.057 & 21.419 & 0.946 & 0.067 & 25.215 & 0.763 & 0.389 & 23.091 & 0.742 & 0.401 \\
NSFF{\cite{li2021neural}}         & - & - & - & - & - & - & 29.365 & 0.829 & 0.278 & 24.163 & 0.795 & 0.289 \\
TiNeuVox{\cite{fang2022fast}}     & 27.988 & 0.960 & 0.063 & 19.612 & 0.940 & 0.073 & 29.982 & 0.864 & 0.213 & 21.069 & 0.909 & 0.281 \\
NVFi{\cite{li2023nvfi}}         & 29.027 & 0.970 & 0.039 & 27.549 & 0.972 & 0.036 & 30.675 & 0.877 & 0.211 & 29.745 & 0.876 & 0.204 \\
DefGS{\cite{yang2024deformable}}        & 37.865 & 0.994 & 0.007 & 19.894 & 0.949 & 0.045 & 30.920 & 0.916 & 0.130 & 21.380 & 0.819 & 0.188 \\
DefGS$_{\text{nvfi}}$ & 37.316 & 0.994 & 0.008 & 28.749 & 0.984 & 0.013 & 29.176 & 0.915 & 0.133 & 31.096 & 0.945 & 0.077 \\
TRACE {\cite{li2025trace}}      & - & - & - & 31.597 & 0.987 & 0.009 & - & - & - & 34.824 & 0.965 & 0.054 \\
FreeGave {\cite{li2025freegave}}  & 39.393 & 0.995 & 0.005 & 31.987 & 0.990 & 0.007 & 32.287 & 0.930 & 0.092 & 35.019 & 0.966 & 0.051 \\
Ours   & \textbf{40.493} & \textbf{0.995} & \textbf{0.005} & \textbf{35.683} & \textbf{0.991} & \textbf{0.006} & \textbf{33.008} & \textbf{0.930} & \textbf{0.091} & \textbf{37.003} & \textbf{0.966} & \textbf{0.050} \\
\bottomrule
\end{tabular}%
}
\end{table*}

Given that the spatial structure is explicitly parameterized by $\bm{\Phi}(\mathbf{x})$, the problem of dynamics learning is effectively reduced to estimating the weighting coefficients of these basis functions as they evolve over time. However, complex dynamic scenes typically exhibit significant \textit{motion heterogeneity}, containing multiple objects with distinct kinematic patterns. Consequently, a single set of global coefficients is evidently insufficient to capture such spatially variant dynamics. To address this limitation, we introduce a latent \textbf{Physics Code} $\mathbf{z}_i \in \mathbb{R}^L$. Extracted via an encoder from the particle's canonical position, this code aims to encapsulate the intrinsic physical attributes of the particle. Subsequently, we employ a lightweight coefficient prediction network $\mathcal{F}_{\text{coef}}$ to accurately regress the time-varying dynamics weights $\mathbf{w}_i(t) \in \mathbb{R}^{12}$, conditioned on the current timestamp $t$ and the specific physics code $\mathbf{z}_i$ as:
\begin{equation}
    \mathbf{w}_i(t) = \mathcal{F}_{\text{coef}}(t, \mathbf{z}_i).
    \label{eq:coef_prediction}
\end{equation}

\begin{figure}
    \centering
    \includegraphics[width=0.5\textwidth,, height=3.5cm]{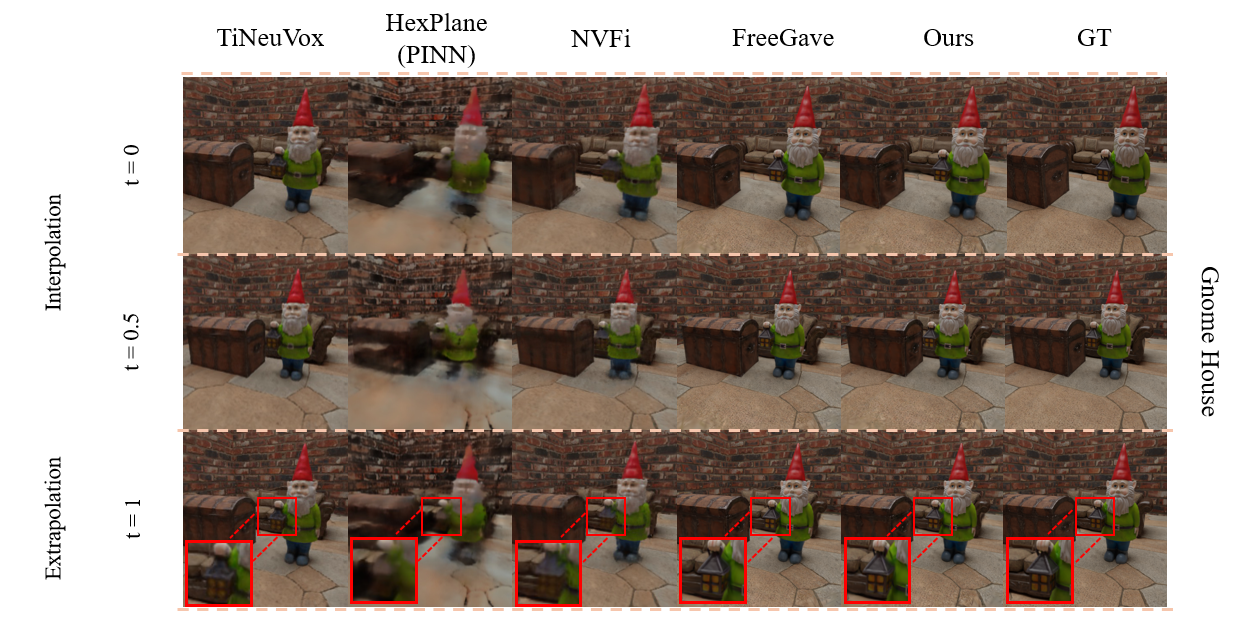}
    \caption{Qualitative comparison of rendering results against other models
    on the Dynamic Indoor Scene Dataset. We adopt the rendering settings from FreeGave\cite{li2025freegave}. }
    \vspace{-0.5cm}
    \label{fig:frame1}
\end{figure}

Although the aforementioned Affine Velocity Model and Physics Code effectively characterize macroscopic motions adhering to physical laws, real-world dynamic scenes frequently exhibit complex, subtle deformations that are ill-suited for description by a strictly affine model. To address this limitation, we introduce a \textbf{Residual Auxiliary Deformation Field}. This module is implemented via a MLP, denoted as $\mathcal{F}_{\text{MLP}}$. It accepts three core inputs: the particle's canonical position $\mathbf{x}_i^{(0)}$, which provides spatial context; the time embedding $\gamma(t)$, indicating the current evolutionary stage; and the Physics Code $\mathbf{z}_i$, which is utilized to modulate deformation patterns based on object-specific local features. Its formula is
\begin{equation}
    (\Delta \mathbf{x}_{\text{t}}, \Delta \mathbf{q}_{\text{t}}, \Delta \mathbf{s}_{\text{t}}) = \mathcal{F}_{\text{MLP}}(\gamma(\mathbf{x}_i^{(0)}), \gamma(t), \mathbf{z}_i).
    \label{eq:auxiliary_net}
\end{equation}

\begin{table}[htbp]
\centering
\caption{Quantitative results for novel view interpolation and future frame extrapolation on the Dynamic Multipart dataset\cite{li2025trace}, rendered at a resolution of 960 × 540.}
\label{tab:freegave_gopro}
\resizebox{\columnwidth}{!}{
\begin{tabular}{l|ccc|ccc}
\toprule
& \multicolumn{6}{c}{Dynamic Multipart Dataset} \\
\cmidrule(lr){2-7}
& \multicolumn{3}{c|}{Interpolation} & \multicolumn{3}{c}{Extrapolation} \\
\cmidrule(lr){2-4} \cmidrule(lr){5-7}
 & PSNR$\uparrow$ & SSIM$\uparrow$ & LPIPS$\downarrow$ & PSNR$\uparrow$ & SSIM$\uparrow$ & LPIPS$\downarrow$ \\
\midrule
T-NeRF{\cite{pumarola2021d}} & - & - & - & 10.064 & 0.576 & 0.537 \\
D-NeRF{\cite{pumarola2021d}} & - & - & - & 13.344 & 0.767 & 0.340 \\
TiNeuVox {\cite{fang2022fast}} & - & - & - & 20.804 & 0.923 & 0.090 \\
NVFi {\cite{li2023nvfi}}     & - & - & - & 25.235 & 0.955 & 0.046 \\
DefGS {\cite{yang2024deformable}}    & 28.061 & 0.942 & 0.055 & 20.664 & 0.930 & 0.067 \\
DefGS$_{\text{nvfi}}$ & - & - & - & 28.455 & 0.979 & 0.017 \\
TRACE {\cite{li2025trace}}      & 35.134 & 0.893 & 0.017 & 33.481 & 0.990 & 0.007  \\
FreeGave {\cite{li2025freegave}}   & 35.631 & 0.990 & 0.016 & 34.667 & 0.990 & 0.007 \\
Ours   & \textbf{35.976} & \textbf{0.990} & \textbf{0.016} & \textbf{34.880} & \textbf{0.992} & \textbf{0.007} \\
\bottomrule
\end{tabular}%
}
\end{table}

The network directly derives three correction quantities relative to the physically deduced state at the current timestamp: displacement correction $\Delta \mathbf{x}_{\text{t}}$, rotation correction $\Delta \mathbf{q}_{\text{t}}$, and scaling correction $\Delta \mathbf{s}_{\text{t}}$. Ultimately, the final state components $\mathbf{s}_i(t)$, $\mathbf{q}_i(t)$, and $\mathbf{x}_i(t)$ of particle $G_i$ at time $t$ are obtained by superimposing the base states $\mathbf{s}_i^{\text{*}}(t)$, $\mathbf{q}_i^{\text{*}}(t)$, and $\mathbf{x}_i^{\text{*}}(t)$ derived from physical integration with their corresponding corrections
\begin{equation}
\left \{
  \begin{aligned}
    \mathbf{x}_i(t) &= \mathbf{x}_i^{*}(t) + \Delta \mathbf{x}_{t}(t) \label{eq:final_pos} \\
    \mathbf{q}_i(t) &= \mathbf{q}_i^{*}(t) \otimes \Delta \mathbf{q}_{t}(t), \\  
    \mathbf{s}_i(t) &= \mathbf{s}_i^{*}(t) \cdot \exp(\Delta \mathbf{s}_{t}(t)) 
   \end{aligned}
   \right.
\end{equation}
where $\mathbf{x}_i^{\text{*}}(t)$, $\mathbf{q}_i^{\text{*}}(t)$, and $\mathbf{s}_i^{\text{*}}(t)$ are derived from the Affine Velocity Model. This design ensures that the dominant motion strictly adheres to physical laws, while simultaneously endowing the model with the flexibility to capture high-frequency geometric details.

\subsection{Global Physical Consistency}
To ensure that the learned dynamics strictly adhere to the laws of fluid mechanics and maintain physical consistency in unobserved regions, we employ \textbf{Global Physical Constraints}. During each training iteration, we randomly sample a batch of \textbf{collocation points} within the spatiotemporal region of interest. Crucially, these points are not confined to the locations of Gaussian particles but span the entire motion space, thereby ensuring the continuity of the velocity field. For each collocation point, we evaluate its velocity field $\mathbf{v}$ and enforce constraints on the network-predicted acceleration $\mathbf{a}$ and velocity $\mathbf{v}$ via the transport equation as

\begin{equation}
    \frac{D\mathbf{v}}{Dt} = \mathbf{a}
    \implies 
    \frac{\partial \mathbf{v}}{\partial t} + \operatorname{grad}(\mathbf{v}) \cdot\mathbf{v} - \mathbf{a} = 0,
    \label{eq:momentum_derivation}
\end{equation}

\begin{equation}
    \frac{\partial \rho}{\partial t} + \nabla \cdot (\rho \mathbf{v}) = 0 
    \xrightarrow[\text{Incompressible}]{\rho = \text{const}} 
    \nabla \cdot \mathbf{v} = 0.
    \label{eq:mass_derivation}
\end{equation}

\section{experiments}
\begin{table*}[htbp]
\centering
\caption{Quantitative results for novel view interpolation and future frame extrapolation on the NVIDIA Dynamic Scene dataset.}
\label{tab:truck_skating}
\resizebox{\linewidth}{!}{
\begin{tabular}{l|ccc|ccc|ccc|ccc}
\toprule
& \multicolumn{6}{c|}{Truck} & \multicolumn{6}{c}{Skating} \\
\cmidrule(lr){2-13}
& \multicolumn{3}{c|}{Interpolation} & \multicolumn{3}{c|}{Extrapolation} & \multicolumn{3}{c|}{Interpolation} & \multicolumn{3}{c}{Extrapolation} \\
\cmidrule(lr){2-4} \cmidrule(lr){5-7} \cmidrule(lr){8-10} \cmidrule(lr){11-13}
Model & PSNR$\uparrow$ & SSIM$\uparrow$ & LPIPS$\downarrow$ & PSNR$\uparrow$ & SSIM$\uparrow$ & LPIPS$\downarrow$ & PSNR$\uparrow$ & SSIM$\uparrow$ & LPIPS$\downarrow$ & PSNR$\uparrow$ & SSIM$\uparrow$ & LPIPS$\downarrow$ \\
\midrule
TiNeuVox{\cite{fang2022fast}}    & 27.230 & 0.846 & 0.229 & 24.887 & 0.848 & 0.209 & \textbf{29.377} & 0.889 & 0.202 & 24.224 & 0.878 & 0.220 \\
HexPlane$_{\text{PINN}}${\cite{cao2023hexplane}} & 25.494 & 0.768 & 0.337 & 24.991 & 0.768 & 0.325 & 24.447 & 0.867 & 0.225 & 23.955 & 0.868 & 0.232 \\
NVFi{\cite{li2023nvfi}}          & 27.276 & 0.840 & 0.235 & 28.269 & 0.855 & 0.220 &26.999 & 0.848 & 0.227 & 28.654 & 0.896 & 0.208 \\
TRACE {\cite{li2025trace}}      & 28.316 & 0.834 & 0.090 & 29.252 & 0.923 & 0.067 & 28.198 & 0.905 & 0.096 & 29.409 & 0.941 & 0.073 \\
FreeGave{\cite{li2025freegave}} & 28.584 & 0.886 & 0.090 & 29.954 & 0.930 & 0.067 & 26.589 & 0.899 & 0.106 & 28.391 & 0.935 & 0.076 \\
Ours  & \textbf{28.924} & \textbf{0.886} & \textbf{0.090} & \textbf{30.304} & \textbf{0.931} & \textbf{0.064} & 28.583 & \textbf{0.902} &  \textbf{0.010} & \textbf{28.665} & \textbf{0.935} & \textbf{0.080} \\
\bottomrule
\end{tabular}%
}
\end{table*}

\textbf{Datasets:} Our method aims to extract underlying physical information from 3D dynamic videos, thereby empowering the model with Future Frame Extrapolation capabilities beyond the observed time horizon, rather than being confined to traditional View Interpolation. To comprehensively evaluate the performance of our method across diverse scenarios, we conduct experiments on the following four datasets:1) \textbf{Dynamic Object Dataset}: Consisting of 6 dynamic objects that encompass a wide range of unique motion patterns, from simple rigid body motions to complex non-rigid deformations.
2) \textbf{Dynamic Indoor Scene Dataset}: Comprising 4 complex indoor scenes. Each scene consists of multiple objects exhibiting distinct trajectories and physical properties, presenting high scene complexity.
3) \textbf{NVIDIA Dynamic Scene Dataset}: Containing two highly challenging real-world dynamic 3D scenes, used to verify the model's generalization capabilities on real-world data.
4) \textbf{Dynamic Multipart Dataset}: Including 4 complex multi-part objects, where distinct parts exhibit diverse motion patterns.

\textbf{Comparisons:}
To comprehensively evaluate the performance of VeloGauss, we select a representative set of methods spanning diverse technical paradigms. As the most relevant physics-aware approach, \textbf{1) NVFi}~\cite{li2023nvfi} incorporates physical constraints but relies entirely on implicit PINN losses to learn priors, whereas we dedicate our effort to constructing a structured dynamics system specifically for explicit Gaussian particles. Within the NeRF framework, we compare against \textbf{2) D-NeRF}~\cite{pumarola2021d}, \textbf{3) T-NeRF}~\cite{pumarola2021d}, \textbf{4) TiNeuVox}~\cite{fang2022fast}, and \textbf{5) NSFF}~\cite{li2021neural}. However, these methods lack the capability for explicit particulate modeling of 3D geometry and appearance, generally falling short of 3DGS-based approaches in terms of both rendering fidelity and inference speed. Regarding 3DGS-based representations, we select \textbf{6) DefGS}~\cite{yang2024deformable}, which achieves high-quality novel view interpolation via deformation fields but inherently lacks the capacity for physical extrapolation. To bridge this gap, we construct a hybrid baseline, denoted as $\textbf{7) DefGS}_{\textbf{NVFi}}$, by integrating DefGS with the velocity field mechanism from NVFi, thereby enabling both novel view synthesis and extrapolation. Finally, we include \textbf{8) FreeGave}~\cite{li2025freegave} and \textbf{9) TRACE}~\cite{li2025trace}, representing the current state-of-the-art in dynamic 3D Gaussian Splatting, both of which utilize a novel framework combining auxiliary deformation fields with Gaussian velocity fields. To ensure strict fairness in comparison, we align the training settings of FreeGave exactly with those of VeloGauss.

\textbf{Metrics:}
For the tasks of Novel View Interpolation and Future Frame Extrapolation, we report standard metrics for RGB view synthesis, including \textbf{PSNR}, \textbf{SSIM}, and \textbf{LPIPS}.

\subsection{Evaluation for Future Frame Extrapolation}
We evaluate our method across all four datasets, covering the core tasks of Novel View Interpolation and Future Frame Extrapolation. All comparative methods are primarily evaluated on the two standard benchmarks: the Dynamic Object Dataset and the Dynamic Indoor Scene Dataset.

\textbf{Results \& Analysis:}
Drawing from the quantitative results in Table~\ref{tab:table1} , Table~\ref{tab:freegave_gopro} and Table~\ref{tab:truck_skating} , our VeloGauss achieves state-of-the-art performance across all evaluated datasets. Most notably, in the highly challenging task of Future Frame Extrapolation, our method significantly outperforms all comparative methods, surpassing recent state-of-the-art approaches such as FreeGave and TRACE by a substantial margin. This compelling evidence confirms that our proposed Particle Dynamics System effectively captures underlying physical information, thereby enabling accurate future motion prediction governed by physical laws. Furthermore, we achieve superior performance compared to other methods in the Novel View Interpolation task. We attribute this comprehensive performance gain to the deep synergy between physical learning and the Auxiliary Deformation Field. This mechanism allows the model to fully leverage Gaussian primitives for the precise representation of complex 3D geometry and fine-grained appearance, all while maintaining physical consistency. As visualized in Figure~\ref{fig:frame1}, VeloGauss delivers accurate and physically plausible predictions even when confronted with complex dynamic multi-object scenes.

\subsection{Ablation Study}
We conduct the following series of ablation studies to validate the effectiveness of the individual components of our method. All ablation experiments are performed on the Dynamic Multipart dataset and the results are summarized in  Table~\ref{tab:ablation}.

\begin{table}[htbp]
\centering
\caption{Quantitative results of the ablation study on the Dynamic Multipart dataset. Here, "K" denotes the number of motion patterns.}
\label{tab:ablation}
\setlength{\tabcolsep}{4pt}
\resizebox{\columnwidth}{!}{%
\begin{tabular}{ccccc|ccc}
\hline
\multirow{2}{*}{} & \multirow{2}{*}{PDS} & \multirow{2}{*}{ADF} & \multirow{2}{*}{GPC} & \multirow{2}{*}{\( K \)} & \multicolumn{3}{c}{Extrapolation} \\
\cline{6-8}
& & & & & PSNR$\uparrow$ & SSIM$\uparrow$ & LPIPS$\downarrow$ \\
\hline
(1) & $\checkmark$ &  & $\checkmark$ & 32 & 9.836 & 0.542 & 0.498 \\
(2) & $\checkmark$ &  &  & 32 & 13.597 & 0.833 & 0.207 \\
(3) &  & $\checkmark$ &  &  32 & 29.016 & 0.980 &0.018 \\
(4) &  & $\checkmark$ & $\checkmark$ & 32 & 32.154 & 0.986 & 0.013 \\
(5) & $\checkmark$ & $\checkmark$ &  & 32 & 33.367 & 0.987 & 0.013 \\
\hline
VeloGauss(Ours) & $\checkmark$ & $\checkmark$ & $\checkmark$ & 32 & \textbf{34.880} & \textbf{0.992} & \textbf{0.007} \\
\hline
\vspace{-0.5cm}
\end{tabular}%
}
\end{table}

 \textbf{Remove Particle Dynamics System (PDS):} We replace the core PDS module with a generic Multilayer Perceptron (MLP). This MLP takes the state of the previous frame as input and directly regresses an unconstrained per-point velocity field $\mathbf{v}(\mathbf{x}, t)$.

\textbf{Remove Auxiliary Deformation Field (ADF):} We remove the ADF module $\mathcal{F}_{MLP}$ and rely solely on the velocity field integrated via RK2 to drive the motion of Gaussian particles.

\textbf{Remove Global Physical Constraints (GPC):} We remove the global physical loss based on PINN optimization during the training process, retaining only the rendering reconstruction loss.

\section{conclusion}
In this paper, we propose VeloGauss, a novel framework built upon 3DGS, designed to jointly learn the geometric structure and latent physical information of dynamic scenes solely from multi-view video inputs, without requiring any manual physical priors or annotations. The core innovation of our approach lies in the integration of a Particle Dynamics System and a Global Physical Constraint mechanism, which enables the precise learning of motion patterns for Gaussian particles. Extensive experiments across four datasets demonstrate that VeloGauss achieves superior performance in both Novel View Interpolation and Future Frame Prediction tasks.
Our method offers significant reference value for enabling applications such as safe autonomous driving, precise robotic manipulation, and high-fidelity virtual reality interactions.

\bibliographystyle{IEEEbib}
\bibliography{icme2025references}

\end{document}